\documentclass[a4paper]{svproc}
\usepackage{url}

\usepackage{hyperref}
\usepackage{subcaption}
\usepackage{tabularx}
\newcolumntype{Y}{>{\centering\arraybackslash}X} 

\usepackage{comment}
\usepackage[binary-units]{siunitx}
\usepackage{relsize}
\usepackage{ifthen}
\usepackage[colorinlistoftodos]{todonotes}

\usepackage[vlined,ruled,linesnumbered]{algorithm2e}
\usepackage{graphics} 
\usepackage{rotating}
\usepackage{color}
\usepackage{enumerate}
\usepackage[T1]{fontenc}
\usepackage{psfrag}
\usepackage{epsfig} 
\usepackage{booktabs}
\usepackage{graphicx,url}
\usepackage{multirow}
\usepackage{array}
\usepackage{latexsym}
\usepackage{amsfonts}
\usepackage{amsmath}
\usepackage{amssymb}
\usepackage{xstring}
\usepackage[noend]{algorithmic}
\usepackage{multirow}
\usepackage{xcolor}
\usepackage{prettyref}
\usepackage{flexisym}
\usepackage{bigdelim}
\usepackage{breqn} 
\usepackage{listings}

\usepackage{enumitem}
\usepackage{xspace}
\usepackage{bm}
\graphicspath{{./figures/}}
\usepackage{tikz}
\usetikzlibrary{matrix,calc}

\usepackage{mdwlist}

\makecompactlist{itemize}{stditemize}

\newcommand{\bdmath}{\begin{dmath}}
\newcommand{\edmath}{\end{dmath}}
\newcommand{\beq}{\begin{equation}}
\newcommand{\eeq}{\end{equation}}
\newcommand{\bdm}{\begin{displaymath}}
\newcommand{\edm}{\end{displaymath}}
\newcommand{\bea}{\begin{eqnarray}}
\newcommand{\eea}{\end{eqnarray}}
\newcommand{\beal}{\beq \begin{array}{ll}}
\newcommand{\eeal}{\end{array} \eeq}
\newcommand{\beas}{\begin{eqnarray*}}
\newcommand{\eeas}{\end{eqnarray*}}
\newcommand{\ba}{\begin{array}}
\newcommand{\ea}{\end{array}}
\newcommand{\bit}{\begin{itemize}}
\newcommand{\eit}{\end{itemize}}
\newcommand{\ben}{\begin{enumerate}}
\newcommand{\een}{\end{enumerate}}

\newcommand{\etal}{\emph{et~al.}\xspace}
\newcommand{\setal}{~\emph{et~al.}\xspace}
\newcommand{\eg}{\emph{e.g.,}\xspace}
\newcommand{\ie}{\emph{i.e.,}\xspace}
\newcommand{\myParagraph}[1]{{\bf #1.}\xspace}

\newcommand{\hide}[1]{}

\newcommand{\hiddenText}{{\color{gray} hidden text.}}
\newcommand{\hideWithText}[1]{\hiddenText}

\newcommand{\scenario}[1]{{\smaller \sf#1}\xspace}

\newcommand{\blue}[1]{{\color{blue}#1}}

\newcommand{\linkToPdf}[1]{\href{#1}{\blue{(pdf)}}}
\newcommand{\linkToPpt}[1]{\href{#1}{\blue{(ppt)}}}
\newcommand{\linkToCode}[1]{\href{#1}{\blue{(code)}}}
\newcommand{\linkToWeb}[1]{\href{#1}{\blue{(web)}}}
\newcommand{\linkToVideo}[1]{\href{#1}{\blue{(video)}}}
\newcommand{\linkToMedia}[1]{\href{#1}{\blue{(media)}}}
\newcommand{\award}[1]{\xspace} 

\newcommand{\ransac}{\scenario{RANSAC}}
\newcommand{\GNC}{\scenario{GNC}}

\newcommand{\pcm}{\scenario{PCM}}

\newcommand{\vislam}{VI-SLAM\xspace}

\newcommand{\bmat}{\left[ \begin{array}}
\newcommand{\emat}{\end{array} \right]}

\newcommand{\bal}{\begin{align}}
\newcommand{\eal}{\end{align}}

\begin{document}
\mainmatter              
\title{Kimera2: Robust and Accurate \\  Metric-Semantic SLAM in the Real World}
\titlerunning{Kimera2: Enhancing Kimera}  
%
\author{
  Marcus Abate\inst{1} \and 
  Yun Chang\inst{1} \and
  Nathan Hughes\inst{1} \and
  Luca Carlone\inst{1}
}
\authorrunning{Marcus Abate et al.} 
%
\tocauthor{
  Marcus Abate, 
  Yun Change, 
  Luca Carlone
}
\institute{
  Laboratory for Information \& Decision Systems (LIDS) \\
  Massachusetts Institute of Technology, Cambridge, USA,\\
    \email{\{mabate,yunchang,lcarlone\}@mit.edu}
}

\maketitle              
\vspace{-10mm}

\begin{tikzpicture}[overlay, remember picture]
\path (current page.north east) ++(-5.3,-0.4) node[below left] {
Accepted for publication at ISER 2023, please cite as follows:
};
\end{tikzpicture}
\begin{tikzpicture}[overlay, remember picture]
\path (current page.north east) ++(-6.9,-0.8) node[below left] {
  M Abate, Y Chang, N Hughes, L Carlone
};
\end{tikzpicture}
\begin{tikzpicture}[overlay, remember picture]
\path (current page.north east) ++(-4.1,-1.2) node[below left] {
  ``Kimera2: Robust and Accurate Metric-Semantic SLAM in the Real World'',
};
\end{tikzpicture}
\begin{tikzpicture}[overlay, remember picture]
\path (current page.north east) ++(-6.5,-1.6) node[below left] {
  IEEE Int. Conf. Exp. Robotics. (ISER), 2023.
};
\end{tikzpicture}

\begin{tikzpicture}[overlay, remember picture]
\path (current page.north west) ++(+16.3,-23.8) node[below left] {
  This work was partially funded by ARL DCIST CRA W911NF-17-2-0181,
};
\end{tikzpicture}
\begin{tikzpicture}[overlay, remember picture]
  \path (current page.north west) ++(+13.0,-24.2) node[below left] {
    by MathWorks, and by an Amazon Research Award.
  };
  \end{tikzpicture}


\vspace{-8mm}
\begin{abstract}

We present improvements to Kimera, an open-source metric-semantic visual-inertial SLAM library. 
In particular, we enhance Kimera-VIO, the visual-inertial odometry pipeline powering Kimera, to support better feature tracking, more efficient keyframe selection, and various input modalities (\eg monocular, stereo, and RGB-D images, as well as wheel odometry). 
Additionally, Kimera-RPGO and Kimera-PGMO, Kimera's pose-graph optimization backends, are updated to support modern outlier rejection methods ---specifically, Graduated-Non-Convexity--- for improved robustness to spurious loop closures. 
These new features are evaluated extensively on a variety of simulated and real robotic platforms, including drones, quadrupeds, wheeled robots, and simulated self-driving cars. 
We present comparisons against several state-of-the-art visual-inertial SLAM pipelines and discuss  strengths and weaknesses of the new release of Kimera. 
The newly added features have been released open-source at \url{https://github.com/MIT-SPARK/Kimera}.

\end{abstract}

\vspace{-10mm}

\begin{figure}[b!]
	\vspace{-4mm}
	\centering
    \begin{subfigure}[b]{0.33\columnwidth}
		\centering
		\includegraphics[width=0.51\columnwidth]{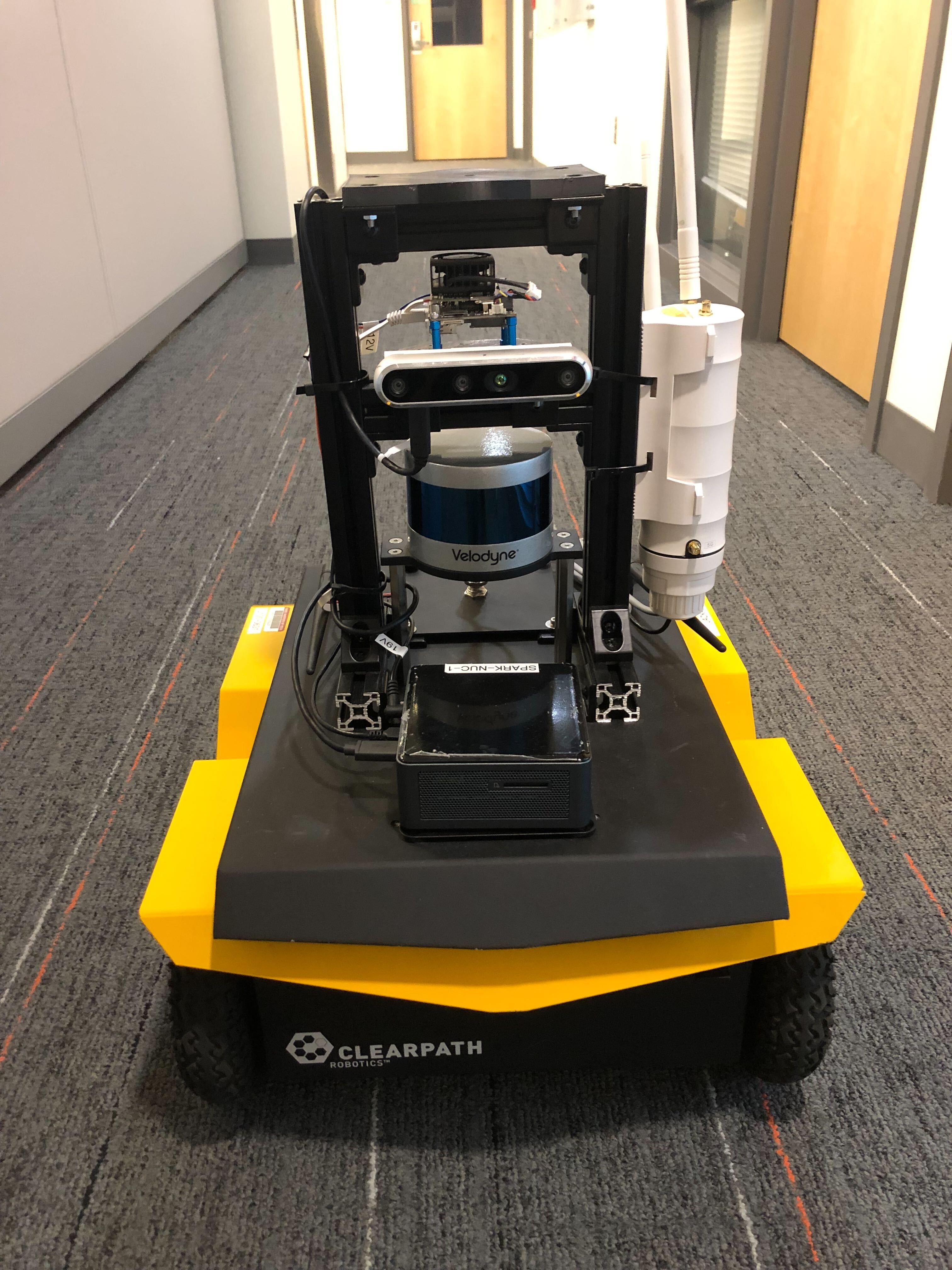}%
		\includegraphics[width=0.51\columnwidth]{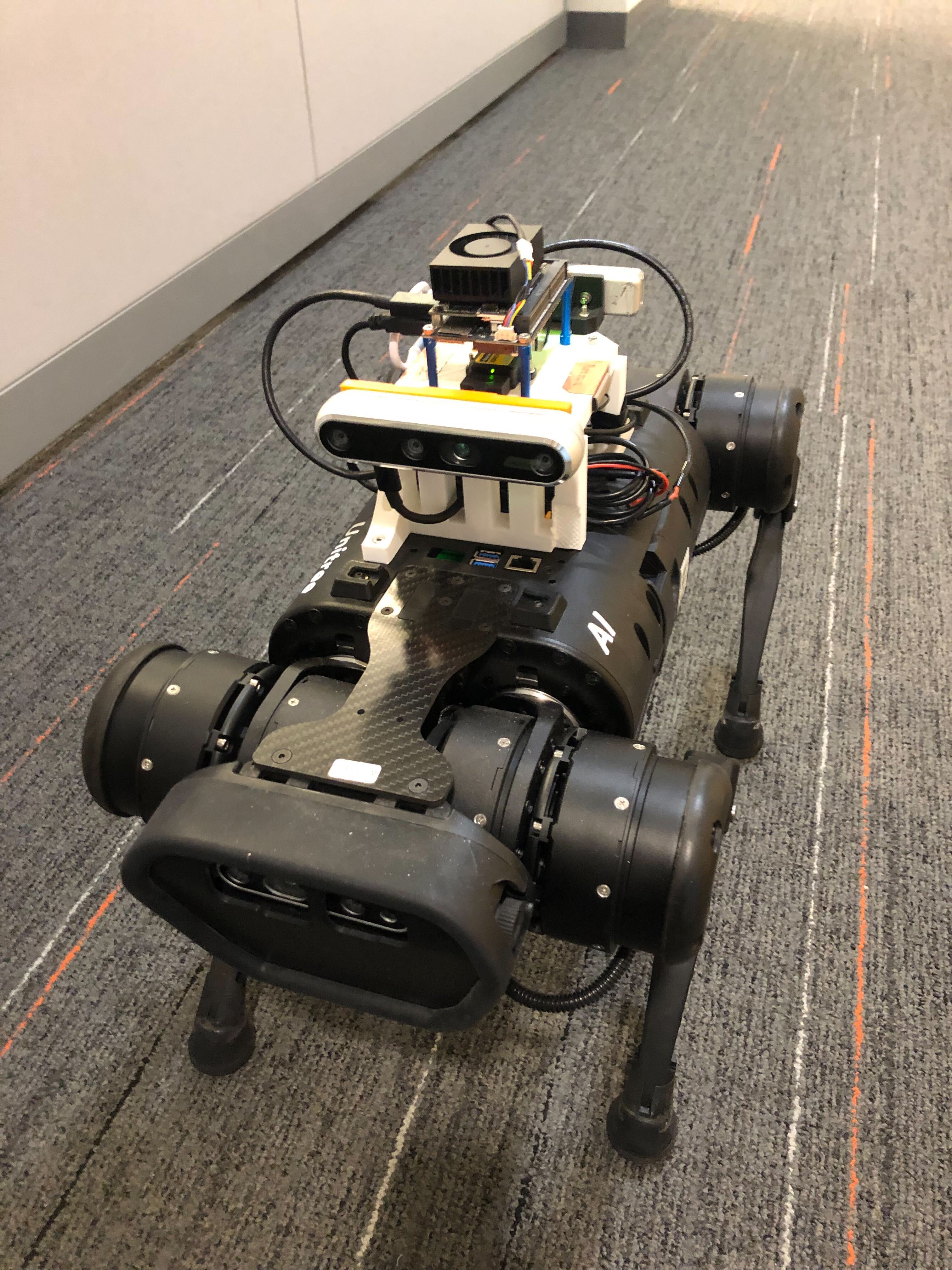}
		\caption{}
		\label{fig:dataset-real}
	\end{subfigure}%
    \begin{subfigure}[b]{0.33\columnwidth}
		\centering
        \includegraphics[width=\columnwidth]{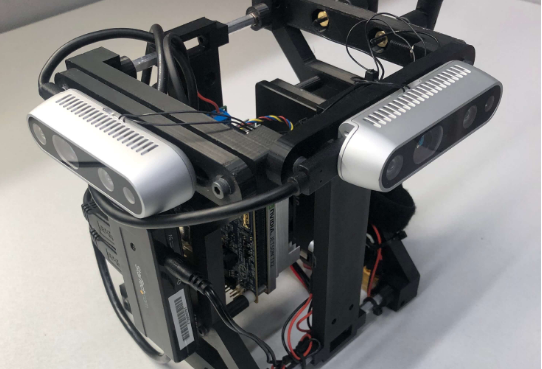}
		\caption{}
		\label{fig:dataset-handheld}
	\end{subfigure}
	\begin{subfigure}[b]{0.33\columnwidth}
		\centering
		\includegraphics[width=0.97\columnwidth]{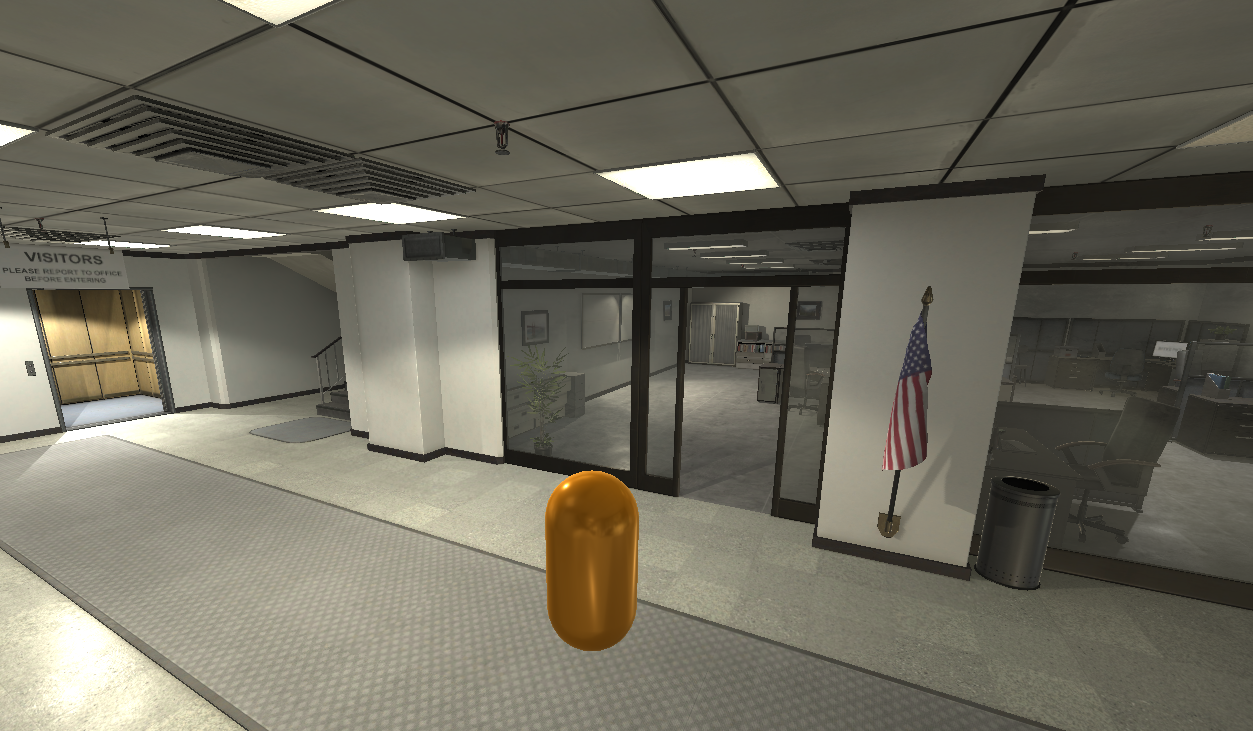}
		\caption{}
		\label{fig:dataset-simSamples1}
	\end{subfigure}%
	\begin{subfigure}[b]{0.33\columnwidth}
		\centering
        \includegraphics[width=0.92\columnwidth]{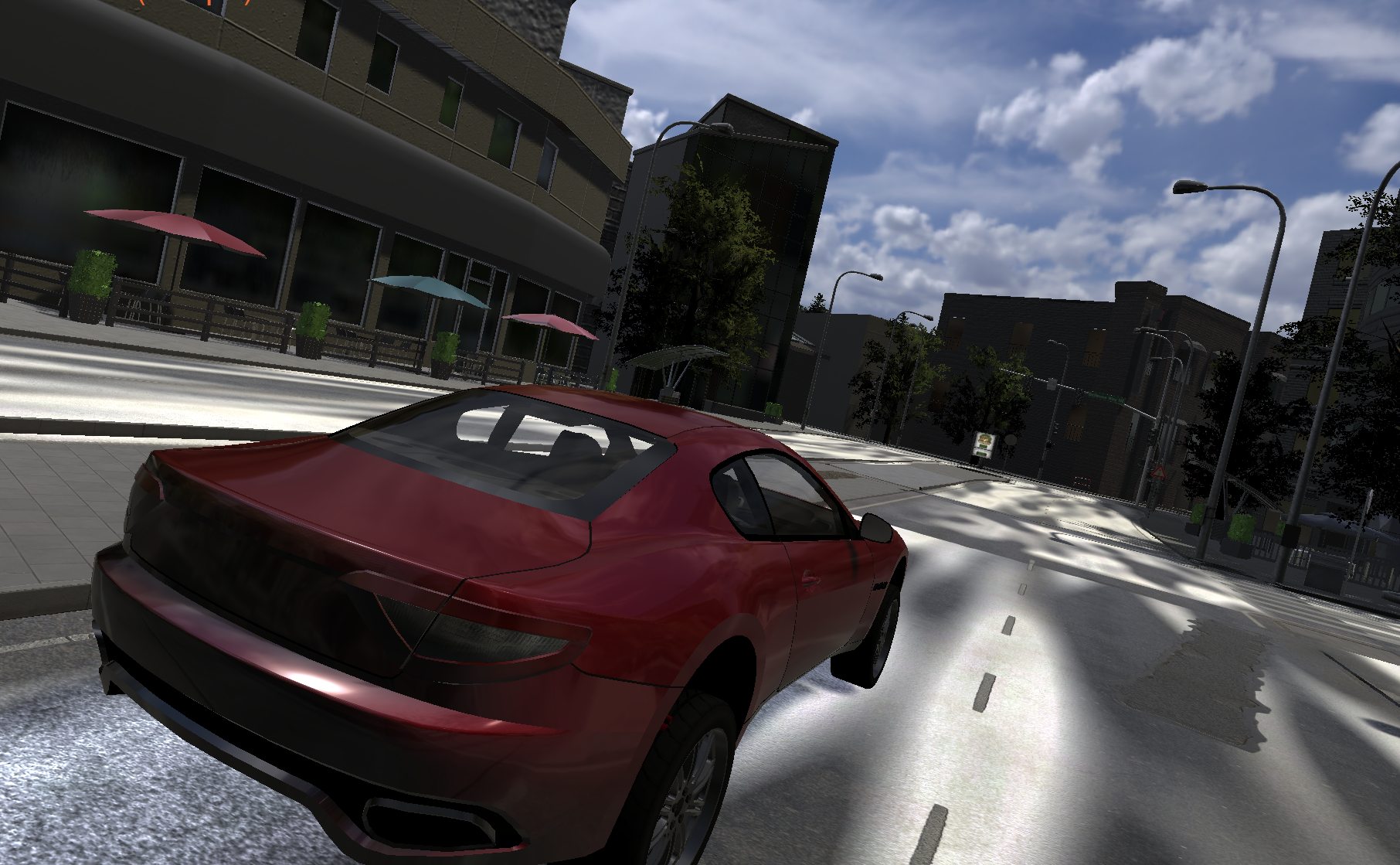}
		\caption{}
		\label{fig:dataset-simSamples2}
	\end{subfigure}%
	\begin{subfigure}[b]{0.33\columnwidth}
		\centering
        \includegraphics[width=0.92\columnwidth]{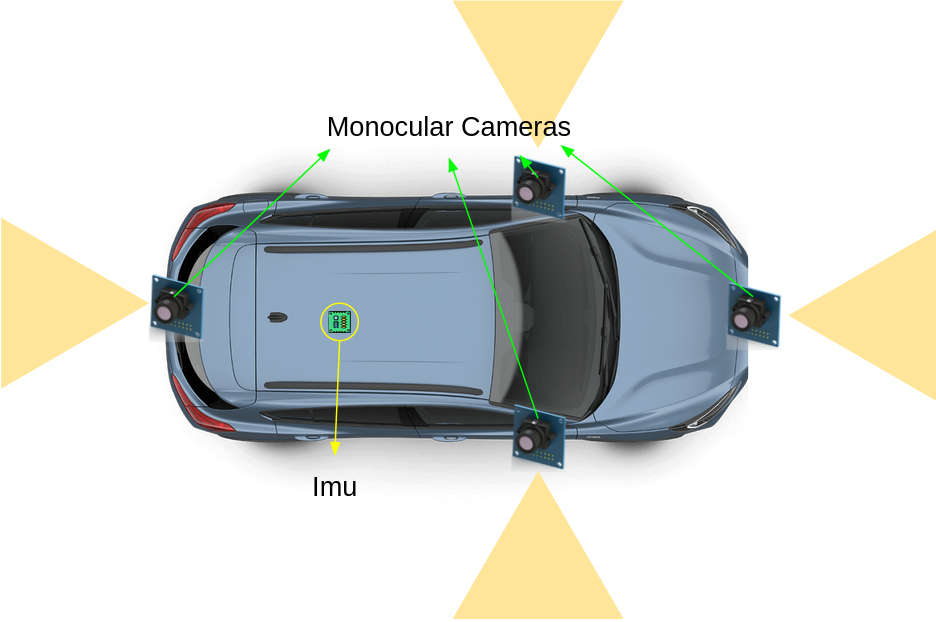}
		\caption{}
		\label{fig:dataset-selfDriving}
	\end{subfigure}
	\caption{
        Overview of some of the platforms and datasets used in the experimental evaluation of Kimera. 
        \textbf{(a)} Clearpath Jackal Robot (left), Unitree A1 quadruped (right). 
        \textbf{(b)} Handheld Jetson-based sensor rig, not discussed in this paper, but evaluated in~\cite{Rosinol21ijrr-Kimera}. 
		\textbf{(c)} uHumans2 simulator office scene. 
		\textbf{(d)} CarSim simulator scene. 
		\textbf{(e)} Self-driving car, not discussed in this paper, but evaluated in~\cite{Abate23iser-KimeraSelfDriving}. 
    }
	\label{fig:all-datasets}
\end{figure}

\section{Introduction}
\label{sec:introduction}

Kimera~\cite{Rosinol20icra-Kimera,Rosinol21ijrr-Kimera} is an open-source metric-semantic visual-inertial SLAM library, released under a permissive BSD license for use by the broader research community and industry.
Since its initial release in 2019, Kimera has been used in several academic and industrial projects. 
End-users of visual-inertial (VI) SLAM pipelines may have diverse system requirements, but generally desire fast (online) performance, as well as accurate and robust state estimation and mapping. 
To meet these performance goals, several new features were implemented in Kimera to improve VIO tracking performance, robust pose graph optimization, and semantic-mapping. 
Additionally, Kimera-Multi~\cite{Chang21icra-KimeraMulti,Tian22tro-KimeraMulti} and Hydra~\cite{Hughes22rss-hydra} made improvements to Kimera-VIO's tracking to serve as a baseline \vislam pipeline for multi-robot mapping and 3D Scene-Graph creation respectively. 

In this paper, we discuss several improvements we made to the open-source version of Kimera, 
that will be released to the public as a part of a version-update to various Kimera packages, 
    including key modules such as Kimera-VIO, Kimera-Semantics, Kimera-RPGO, and Kimera-PGMO. 
We showcase ablation studies on select features added to Kimera since its release in 2019, and perform comparisons against the state of the art (in particular, ORB-SLAM3 and Vins-Fusion). 
We evaluate Kimera's strengths and weaknesses as well as its versatility across various platforms. 
Experiments are conducted on datasets gathered from real-life platforms as well as simulated environments (Fig.~\ref{fig:all-datasets}). 

This paper is organized as follows: Section~\ref{sec:relatedWork} covers related work. 
Section~\ref{sec:systemarchitecture} provides a detailed explanation of selected features added to Kimera that will be evaluated. 
Section~\ref{sec:experiments} describes ablation tests and comparisons against the state of the art, and discusses the results of our experiments. 
Finally, Section~\ref{sec:conclusions} concludes the paper and provides avenues of future research and development. 


\section{Related Work}
\label{sec:relatedWork}

\myParagraph{\vislam} There are several open-source \vislam algorithms that have reached maturity for research and industrial applications. 
Kimera~\cite{Rosinol21ijrr-Kimera,Rosinol20icra-Kimera}, Vins-Fusion~\cite{Qin19arxiv-VINS-Fusion-odometry}, ORB-SLAM3~\cite{Campos21-TRO}, and Open-Vins~\cite{Geneva20icra-openVINS} are some of the most up-to-date packages used for robotics. 
All leverage modern sensing systems, in particular stereo or RGB-D cameras with IMU sensors, which generally enable sub-meter trajectory estimation and 3D mapping accuracy over long distances in challenging visual environments. 
While all these pipelines are capable of performing real-time trajectory estimation, to our knowledge Kimera is still the only open-source pipeline that performs online \vislam as well as real-time metric-semantic dense 3D reconstruction. 
The interested reader can find a more in-depth survey of the \vislam literature in our previous work~\cite{Rosinol21ijrr-Kimera}. 

\myParagraph{Metric-semantic SLAM} 
In the years since Kimera's release, we have seen novel approaches for robot spatial perception. 
Hughes\setal~\cite{Hughes23arxiv-hydraFoundations} build on Kimera and develop a real-time approach for  3D scene graph construction.
Jatavallabhula\setal~\cite{Jatavallabhula23arxiv-conceptFusion} introduce \emph{ConceptFusion}, a system that generates an open-set dense map of the environment, 
 which is queryable by text, image, audio, and clicks. 
The authors extended the work to create 3D scene graphs in \emph{ConceptGraphs}~\cite{Kuwajaerwala23arxiv-conceptGraphs}. 
Other recent efforts to leverage scene graph representations include Tourani\setal~\cite{Tourani23arxiv-optimizableSceneGraph}, who improve state estimates with fiducial markers incorporated into the scene graph's topology. 
Other systems leverage large, internet-scale trained models, including large language models (LLMs).
Kassab\setal~\cite{Kassab23arxiv-lexis} use LLMs in conjunction with VIO to gather semantic information about a scene and perform loop closures using CLIP features. 
More broadly, LLMs have been applied to a variety of perception tasks, from segmentation to searching a 3D map~\cite{Kerr23iccv-lerf,Shafiullah23arxiv-clipFields}. We refer the reader to~\cite{Hughes23arxiv-hydraFoundations} for a more extensive review.


\section{Kimera2: System Improvements}
\label{sec:systemarchitecture}

\subsection{Kimera-VIO Frontend}
\label{subsec:features-frontend}

Kimera-VIO's frontend serves as an initial data-processing module to prepare raw sensor measurements for optimization in the VIO backend. 
The frontend is flexible enough to be implemented for a variety of sensor inputs. 
In the original version of Kimera-VIO~\cite{Rosinol20icra-Kimera,Rosinol21ijrr-Kimera} the frontend was implemented for stereo cameras and IMU, assumed synchronized. 
Kimera was designed from the ground up to be highly modular and developer-friendly, to support expansion in the future. 
This allowed us to add implementations for monocular-IMU input, RGB-D cameras with IMU, and external (\eg wheel) odometry, which we have since released to the community. 
Below we provide details for a subset of the newly added features to Kimera-VIO's frontend module. 

\myParagraph{External odometry}
Various robotics platforms provide alternative odometry estimates. 
For instance, platforms with onboard LiDAR sensors can provide LIDAR odometry; 
platforms on wheels give easy access to wheel odometry; some cameras, such as the RealSense T265, also provide an off-the-shelf odometry estimate. 
Fusing these various inputs with Kimera-VIO's estimate without implementing sensor-specific frontends serves as a fast way to improve the state estimate without slowing down Kimera's pose estimation thread. 
Therefore, we optionally process external odometry as relative poses between VIO keyframes in a separate submodule of the frontend. 
These relative poses are passed to the VIO backend alongside visual features and preintegrated IMU measurements. 
The VIO backend then combines the visual features (stereo, mono, or RGB-D), IMU measurements, and odometric measurements (modeled as {\tt BetweenFactor} in GTSAM) to compute an improved odometry estimate over a receding horizon. 

\myParagraph{Feature binning and non-max suppression}
For visual inputs, we implement two small improvements that enable more efficient processing of images and keypoint tracking. 
Feature binning allows the user to provide an abstracted pixel-mask of the image defining which portions of the image to include in feature detection and which to ignore; it also allows more uniform detection of features across the unmasked portion of the image.
This is most useful in situations where parts of the image are expected to be unusable; for example, if fisheye cameras are used onboard a self-driving car, it is possible that the chassis is constantly visible in the camera field of view, potentially hindering VIO performance.
In addition, we implement various flavors of non-maximum suppression~\cite{Bailo18pami-nms}, 
that allows users to aggressively cull feature tracks from the frontend, improving feature tracking and focusing computation on the most promising features.

\myParagraph{Keyframe logic}
At the time a keyframe is identified, all frontend measurements (including visual features and pre-integrated IMU measurements and any other optional data) between the previous keyframe and the current keyframe are sent to the backend for inclusion in the VIO fixed-lag smoother. 
By restricting optimization to only keyframes, we can include more visual measurements in the factor graph without slowing down the optimization thread. 
Kimera's previous logic for choosing keyframes was based on a user-defined parameter which determined the elapsed time between keyframes. 
However, we found that for vehicles such as cars, trajectories had long periods of minimal-to-zero movement. 
During these times, choosing keyframes and triggering backend optimizations at a constant rate was unnecessary as the pose had not deviated significantly from the previous keyframe, and could lead to increased drift or even failures. 
For this reason we modified the keyframe logic to choose keyframes either when a maximum amount of time had passed since the previous keyframe, or when there was sufficient disparity between keyframes (in terms of optical flow of the features) to warrant a new keyframe. 
The latter condition pushes the frontend to only select keyframes after the robot had moved, saving computation. 
Additionally, because the backend factor graph operates over a receding horizon (\ie it is a fixed-lag smoother), by choosing keyframes only after the robot has moved we prevent Kimera from ``forgetting'' the entire recent trajectory prior to the robot standing still, thus avoiding degenerate conditions. 
The updated keyframe logic generally leads to smaller factor-graph sizes while retaining enough information about the past to maintain tracking during longer periods of minimal movement. 

\subsection{GNC for Kimera-RPGO and Kimera-PGMO}
\label{subsec:features-backend}


Kimera-VIO's backend creates and optimizes a factor graph of various measurements collected from the frontend over a receding time horizon, to estimate the robot odometry. 
This odometry estimate is then combined with loop closure detections to compute a globally consistent trajectory estimate.
In particular,  
Kimera-LCD (Loop-Closure Detection) processes backend odometry in conjunction with frontend data (\eg images) associated with each keyframe in order to identify loop closures, using a visual-Bag-of-Words approach~\cite{Galvez12tro-dbow}. 
Both odometry factors and loop-closure factors are then added to a separate pose-graph which is optimized using Kimera-RPGO (Robust Pose Graph Optimization)~\cite{Rosinol20icra-Kimera,Rosinol21ijrr-Kimera}. 

In the past, RPGO relied on an incremental version of Pairwise Consistent Measurement Set Maximization (\pcm)~\cite{Mangelson18icra} for rejection of spurious loop closures. 
This enabled the rejection of bad loop-closure candidates, which can be frequent when using the visual-Bag-of-Words method in scenes where the environment is visually similar in many areas. 
In~\cite{Rosinol21ijrr-Kimera}, we showed that Kimera-RPGO with \pcm led to drastic improvements in global pose estimation. 
However, since then newer outlier rejection methods have come to the forefront of the field. 
Yang~\etal~\cite{Yang20ral-GNC} show that Graduated-Non-Convexity (\GNC) leads to superior performance in pose graph optimization. The approach is further validated across multiple applications in~\cite{Antonante21tro-outlierRobustEstimation} and compared against \ransac and \pcm. 
As this is relevant to \vislam, \GNC is now implemented in Kimera-RPGO as an option for outlier rejection on the pose-graph optimization. 
Finally, in~\cite{Rosinol21ijrr-Kimera} we presented Kimera-PGMO for jointly optimizing the pose-graph and the dense metric-semantic mesh. 
\GNC can be used here as well since the underlying optimization framework is shared with Kimera-RPGO, so we have also modified Kimera-PGMO to use \GNC for more accurate mesh reconstruction. 




\section{Experiments}
\label{sec:experiments}

As Kimera is easily adaptable to a variety of robotic platforms, in this section we provide experimental results for Kimera on a diverse array of datasets and validate the new features discussed in Section~\ref{sec:systemarchitecture}.
Additionally, we provide comparisons against other state-of-the-art open-source \vislam pipelines. 

\subsection{Datasets}
\label{subsec:experiments-datasets}

We include results on a wide range of datasets ---most of which are publicly available--- so as to highlight the specific impact of each feature discussed and prove the flexibility of Kimera as a broadly applicable \vislam library. 
Fig.~\ref{fig:all-datasets} provides an overview of the various datasets. 
In addition to the simulated and real data described below, we also evaluate performance on EuRoC sequences~\cite{Burri16ijrr-eurocDataset}. 
These consist of data collected from a drone platform. 

\myParagraph{A1 and Jackal}
Many of these datasets have been collected in the context of the Kimera-Multi~\cite{Chang21icra-KimeraMulti,Tian22tro-KimeraMulti,Tian23iros-KimeraMultiExperiments} project, and include (now publicly available) datasets collected on Unitree A1 quadrupeds and Clearpath Robotics Jackal wheeled robots. 
The A1 is a quadrupedal robot with an onboard RealSense D455 RGB-D Camera for sensing, as well as IMU and external odometry. 
The Jackal is a small four-wheeled ground robot with a stereo camera and IMU, as well as wheel odometry. 
Datasets were recorded in a wide range of locations on MIT's campus, including indoor and outdoor locations, underground tunnels, and an undergraduate dorm. 
Datasets labeled {\tt indoor} are datasets collected with the Jackal robot in indoor environments across MIT's campus. 
Datasets labeled {\tt outdoor} are Jackal datasets collected in outdoor environments across MIT. 
Datasets labeled {\tt hybrid} are Jackal datasets where the robot transitions between indoor and outdoor. 
We also include one A1 dataset ---labeled {\tt a1}--- that was recorded inside an undergraduate dorm hall (Simmons). 


\myParagraph{uHumans2}
The uHumans2 dataset was presented as part of our earlier work on Kimera~\cite{Rosinol21ijrr-Kimera}. 
The simulation environment was open-sourced, as were the datasets. 
The agent has a forward-facing stereo camera and simulated IMU. 
Simulation environments are varied, from a small apartment to a large underground subway station. 

\myParagraph{CarSim}
Datasets labeled {\tt carsim} are collected inside the TESSE environment~\cite{Rosinol20icra-Kimera,Rosinol21ijrr-Kimera,Rosinol20rss-dynamicSceneGraphs}. 
However, unlike in the uHumans2~\cite{Rosinol20icra-Kimera,Rosinol21ijrr-Kimera} dataset, these sequences are recorded in a simulated outdoor urban environment, using a car as the robotic agent. 
The simulated car has four monocular cameras mounted in the front, rear, left, and right. 
For these ablation tests, we use Kimera in monocular-mode with the right camera. 
For more results on self-driving platforms, refer to~\cite{Abate23iser-KimeraSelfDriving}. 




\subsection{External Odometry}
\label{subsec:experiments-external-odometry}

Wheel odometry was available on the Jackal robot. 
Table~\ref{tab:external_odom_ablation} shows datasets from the Jackal, and the effect the inclusion of wheel odometry had on the localization performance. 
Three trials were performed for each dataset, and the reported metrics are the mean and standard deviation of the RMSE of the Absolute Translation Error (ATE) across all trials. 
The external odometry offers an advantage in many cases, in particular in outdoor datasets. 
However, for some indoor datasets the error was slightly higher with external odometry factors ({\tt jackal_hybrid_3} is predominantly indoors). 
This is likely because visual features are easier to track in these environments, and since they are mostly close to the camera, using a stereo configuration leads to more accurate depth and trajectory estimates in these cases. 
Any error in the wheel odometry therefore has an outsized negative effect, as the vision factors already achieve high accuracy.


\begin{table}[h!]
  \vspace{-5mm}
  \centering
  \footnotesize
  \begin{tabularx}{\columnwidth}{l *{4}{Y}}
    \toprule
    & \multicolumn{4}{c}{Absolute Translation Error RMSE} \\
    \cmidrule{2-5}
    & \multicolumn{2}{c}{\textbf{Without EO}} & \multicolumn{2}{c}{\textbf{With EO}} \\
Dataset & Avg [m] & Std [m] & Avg [m] & Std [m] \\
  \midrule
  jackal_hybrid_0    & 3.25                    & 0.03  & \textcolor{green}{3.21} & 0.04  \\
  jackal_hybrid_1    & 4.1                     & 0.35  & \textcolor{green}{3.73} & 0.39  \\
  jackal_hybrid_2    & --                      & --    & \textcolor{green}{8.3 } & 1.09  \\
  jackal_hybrid_3    & \textcolor{green}{9.67} & 0.67  & 11.8                    & 7.02  \\
  jackal_indoor_0    & \textcolor{green}{9.17} & 1.64  & 11.4                    & 1.19  \\
  jackal_indoor_1    & 3.98                    & 2.0   & \textcolor{green}{3.98} & 0.83  \\
  jackal_indoor_2    & 8.67                    & 3.24  & \textcolor{green}{6.97} & 2.05  \\
  jackal_indoor_3    & 7.06                    & 0.88  & \textcolor{green}{6.04} & 0.81  \\
  jackal_outdoor_0   & 10.6                    & 0.85  & \textcolor{green}{10.6} & 1.99  \\
  jackal_outdoor_1   & 15.9                    & 1.02  & \textcolor{green}{12.3} & 1.35  \\
  jackal_outdoor_2   & \textcolor{green}{17.0} & 3.99  & 21.3                    & 2.09  \\
  \bottomrule
\end{tabularx}%
\caption{
    VIO accuracy with and without external (wheel) odometry. 
    Datasets come from the KimeraMulti~\cite{Chang21icra-KimeraMulti,Tian22tro-KimeraMulti,Tian23iros-KimeraMultiExperiments} project, and were collected with Jackal robots. 
    The best result for each dataset is highlighted in green. A dash is used to denote that Kimera failed to get a reasonable trajectory for that dataset in the given configuration. 
}
\label{tab:external_odom_ablation}
\vspace{-5mm}
\end{table}{}




\subsection{Feature Binning}
\label{subsec:experiments-feature-binning}

Feature binning was performed on the A1 dataset from KimeraMulti, as well as the CarSim datasets; in other datasets it had no effect as the mask was not necessary. 
In the case of the A1, the binning mask was designed to remove features from the body of the robot, visible in the bottom of the camera image. 
For the CarSim datasets, features typically associated with the sky (center and high in the frame) were masked off to improve performance. 
Table~\ref{tab:binning_ablation} shows the results of this ablation study. 
In the A1 case, Kimera failed completely without the binning mask, and this was observed in other datasets recorded on the A1 as well. 
For the CarSim datasets, localization error was better across the board when binning was enabled. 
For applications with known regions of bad features, this method seems to be an effective solution for reducing localization error. 


\begin{table}[h!]
  \vspace{-5mm}
  \centering
  \footnotesize
  \begin{tabularx}{\columnwidth}{l *{4}{Y}}
    \toprule
    & \multicolumn{4}{c}{Absolute Translation Error RMSE} \\
    \cmidrule{2-5}
    & \multicolumn{2}{c}{\textbf{No Binning}} & \multicolumn{2}{c}{\textbf{Binning}} \\
Dataset & Avg [m] & Std [m] & Avg [m] & Std [m] \\
  \midrule
  carsim_1  & 1.22 & 0.5  & \textcolor{green}{0.65}  & 0.03  \\
  carsim_2  & 0.82 & 0.43 & \textcolor{green}{0.51}  & 0.01  \\
  carsim_3  & 3.67 & 0.76 & \textcolor{green}{2.55}  & 0.52  \\
  carsim_4  & 8.52 & 3.5  & \textcolor{green}{3.22}  & 0.26  \\
  \midrule
  a1_simmons_0       & -- & -- & \textcolor{green}{1.74} & 0.33  \\
  \bottomrule
\end{tabularx}%
\caption{
    VIO accuracy with and without feature binning. 
    A dash is used to denote that Kimera failed to get a reasonable trajectory for that dataset in the given configuration. 
}
\label{tab:binning_ablation}
\vspace{-10mm}
\end{table}{}


\subsection{Keyframe Logic}
\label{subsec:experiments-keyframe-logic}

In Table~\ref{tab:mdsl_ablation}, we perform an ablation study on the {\tt max_disparity_since_lkf} parameter in Kimera-VIO's frontend. 
This determines what is the disparity threshold (in terms of average norm of the optical flow of the features)
 past which a new keyframe is selected.
The higher the value, the more the features can move in the frame before a keyframe is selected and the backend factor-graph optimization is triggered. 
When set to {\tt 1000}, the system is essentially disabled, defaulting to the time-based logic of the previous version of Kimera. 
The uHumans2 and CarSim sequences were not included as the agent is at a constant velocity for the majority of each sequence, so the parameter had little effect. 
We see that the best results are generally biased towards smaller values for {\tt max_disparity_since_lkf}, confirming that disparity in optical flow is a superior method for identifying keyframes. 
In some cases, the difference between the best and worst result for each dataset is quite large (by an order of magnitude). 
Overall, selecting a {\tt max_disparity_since_lkf} value between {\tt 75-100} appears to give consistently good results. 


\begin{table*}[!]
  \vspace{-7mm}
  \centering
  \begin{tabularx}{\textwidth}{l *{12}{Y}}
    \toprule
    & \multicolumn{12}{c}{Absolute Translation Error RMSE} \\
    \cmidrule{2-13}
    & \multicolumn{2}{c}{\scriptsize \textbf{MDSL 25}} & \multicolumn{2}{c}{\scriptsize \textbf{MDSL 50}} & \multicolumn{2}{c}{\scriptsize \textbf{MDSL 75}} & \multicolumn{2}{c}{\scriptsize \textbf{MDSL 100}} & \multicolumn{2}{c}{\scriptsize \textbf{MDSL 150}} & \multicolumn{2}{c}{\scriptsize \textbf{MDSL 1000}} \\
Dataset & Avg [m] & Std [m] & Avg [m] & Std [m] & Avg [m] & Std [m] & Avg [m] & Std [m] & Avg [m] & Std [m] & Avg [m] & Std [m] \\
  \midrule
  jackal_hybrid_0  & \textcolor{green}{3.25} & 0.12 & 5.57 & 2.14 & 3.74 & 0.81 & 3.31 & 0.04 & 3.29 & 0.15 & 3.27 & 0.09 \\
  jackal_hybrid_1  & 3.25 & 0.36 & \textcolor{green}{2.78} & 1.48 & 3.56 & 0.12 & 3.41 & 0.54 & 3.56 & 0.66 & 3.71 & 0.35 \\
  jackal_hybrid_2  & 11.7 & 0.43 & --   & --   & \textcolor{green}{9.58} & 2.62 & 11.6 & 1.59 & 17.1 & 4.65 & --   & --   \\
  jackal_hybrid_3  & 70.6 & 3.62 & 54.3 & 25.2 & 37.7 & 42.6 & \textcolor{green}{9.79} & 3.27 & 26.3 & 16.2 & --   & --   \\
  jackal_indoor_0  & \textcolor{green}{1.53} & 2.63 & 7.68 & 0.61 & 7.99 & 0.81 & 7.2  & 1.76 & 6.32 & 1.71 & 8.56 & 1.8  \\
  jackal_indoor_1  & 2.73 & 0.49 & 2.83 & 0.52 & 4.59 & 1.89 & \textcolor{green}{2.67} & 0.31 & 3.21 & 0.69 & 2.91 & 0.06 \\
  jackal_indoor_2  & 13.9 & 8.1  & --   & --   & 9.08 & 1.76 & 8.58 & 3.11 & 9.07 & 3.05 & \textcolor{green}{6.85} & 1.15 \\
  jackal_indoor_3  & 5.2  & 2.56 & 6.36 & 2.27 & \textcolor{green}{4.46} & 0.24 & 6.17 & 2.23 & 7.38 & 3.45 & 4.64 & 1.38 \\
  jackal_outdoor_0 & 4.64 & 1.38 & 5.5  & 0.68 & 5.04 & 0.41 & \textcolor{green}{4.23} & 0.61 & 6.94 & 3.65 & 5.33 & 0.51 \\
  jackal_outdoor_1 & 9.78 & 1.36 & 12.5 & 1.56 & \textcolor{green}{9.35} & 2.36 & 12.4 & 0.61 & 13.2 & 1.05 & 12.4 & 1.68 \\
  jackal_outdoor_2 & \textcolor{green}{10.7} & 2.03 & 16.2 & 3.17 & 15.5 & 2.48 & 15.7 & 0.18 & 11.9 & 1.21 & 16.4 & 2.95 \\
  \midrule
  a1_simmons_0       & 1.13 & 2.21 & 1.66 & 2.33  & \textcolor{green}{0.92} & 0.64  & 1.14                    & 4.1   & 0.99                    & 0.16  & 1.19 & 2.10  \\
  \bottomrule
\end{tabularx}%
\caption{
    VIO accuracy ablation study on keyframe logic for Jackal and A1 datasets. 
    Dashes are used to denote tracking failures (very high error). 
}
\label{tab:mdsl_ablation}
\vspace{-15mm}
\end{table*}{}

\subsection{\GNC vs \pcm}
\label{subsec:experiments-gnc}

We compare \GNC with a baseline based on Pairwise Consistency Maximization (\pcm)~\cite{Mangelson18icra} for robust pose graph optimization. 
On all datasets, the rotation threshold for \pcm was {\tt 0.01} and the translation threshold was {\tt 0.05}. 
There were several loop closure candidates in all of the datasets surveyed. 
Table~\ref{tab:pcm_gnc_ablation} shows that \GNC improves localization performance substantially in the majority of cases. 
Sequences from KimeraMulti and uHumans2 are included. 
In the case of the A1 robot ({\tt a1_simmons_0}), \GNC was required to make Kimera work. 


\begin{table}[h!]
  \vspace{-5mm}
  \centering
  \footnotesize
  \begin{tabularx}{\columnwidth}{l *{4}{Y}}
    \toprule
    & \multicolumn{4}{c}{Absolute Translation Error RMSE} \\
    \cmidrule{2-5}
    & \multicolumn{2}{c}{\textbf{PCM}} & \multicolumn{2}{c}{\textbf{GNC}} \\
Dataset & Avg [m] & Std [m] & Avg [m] & Std [m] \\
  \midrule
  jackal_hybrid_0    & 3.28                    & 0.08  & \textcolor{green}{3.23} & 0.03 \\
  jackal_hybrid_1    & \textcolor{green}{3.79} & 0.16  & 3.91                    & 0.33 \\
  jackal_hybrid_2    & 14.6                    & 5.34  & \textcolor{green}{12.3} & 3.78 \\
  jackal_hybrid_3    & 14.7                    & 3.39  & \textcolor{green}{11.2} & 6.62 \\
  jackal_indoor_0    & 9.84                    & 3.03  & \textcolor{green}{8.02} & 1.19 \\
  jackal_indoor_1    & \textcolor{green}{3.41} & 0.9   & 3.99                    & 1.11 \\
  jackal_indoor_2    & 6.39                    & 0.96  & \textcolor{green}{6.13} & 0.73 \\
  jackal_indoor_3    & 7.55                    & 3.28  & \textcolor{green}{6.33} & 2.53 \\
  jackal_outdoor_0   & 12.4                    & 1.75  & \textcolor{green}{11.2} & 0.69 \\
  jackal_outdoor_1   & \textcolor{green}{13.8} & 6.52  & 16.2                    & 0.83 \\
  jackal_outdoor_2   & 16.4                    & 4.06  & \textcolor{green}{14.0} & 4.14 \\
  \midrule
  a1_simmons_0          & --                      & --    & \textcolor{green}{28.5} & 2.43 \\
  \midrule
  uH2_apartment     & 0.1  & 0.6     & \textcolor{green}{0.1}  & 0.0  \\
  uH2_suburb        & --   & --      & \textcolor{green}{2.47} & 0.67 \\
  uH2_office        & 0.33 & 0.4     & \textcolor{green}{0.33} & 0.09 \\
  uH2_subway        & 4.97 & 0.91    & \textcolor{green}{4.11} & 1.43 \\
  \bottomrule
\end{tabularx}%
\caption{
    \vislam accuracy using PCM and GNC for loop closure outlier rejection, on real and simulated data. 
}
\label{tab:pcm_gnc_ablation}
\vspace{-10mm}
\end{table}{}


\subsection{PGMO}
\label{subsec:experiments-pgmo}

In~\cite{Rosinol21ijrr-Kimera}, we showed the effect of Kimera-PGMO on mesh reconstruction. 
We found that by jointly optimizing the pose-graph with loop closures and the mesh, we were able to close loops on the dense metric-semantic mesh and obtain a higher accuracy in the mesh. 
Table~\ref{tab:pgmo_ablation} compares Kimera-Semantics with Kimera-PGMO, where Kimera-Semantics is the original version released in~\cite{Rosinol20icra-Kimera}. 
Kimera-PGMO provides better mesh reconstruction accuracy due to the inclusion of loop closure factors. 
This evaluation was done in our previous work~\cite{Rosinol21ijrr-Kimera} on the uHumans2 dataset, however we replicate the experiment here on new data (CarSim and the Kimera-Multi datasets). 


\begin{table}[h!]
  \vspace{-5mm}
    \centering
    \footnotesize
    \begin{tabularx}{\columnwidth}{l *{4}{Y}}
      \toprule
      & \multicolumn{4}{c}{3D Mesh Geometric Reconstruction RMSE} \\
      \cmidrule{2-5}
      & \multicolumn{2}{c}{\textbf{Kimera-Semantics}} & \multicolumn{2}{c}{\textbf{Kimera-PGMO}} \\
  Dataset & Avg [m] & Std [m] & Avg [m] & Std [m] \\
    \midrule
    carsim_1 & 0.26                     & 0.05   & \textcolor{green}{0.22}  & 0.06  \\
    carsim_2 & 0.40                     & 0.1    & \textcolor{green}{0.37}  & 0.05  \\
    carsim_3 & 0.32                     & 0.11   & \textcolor{green}{0.29}  & 0.07  \\
    carsim_4 & 0.35                     & 0.09   & \textcolor{green}{0.34}  & 0.12  \\
    \midrule
    jackal_hybrid_0   & 0.76                     & 0.01   & \textcolor{green}{0.67 } & 0.01  \\
    jackal_hybrid_1   & 1.59                     & 0.16   & \textcolor{green}{1.42 } & 0.09  \\
    jackal_hybrid_2   & 1.61                     & 0.21   & \textcolor{green}{1.61 } & 0.21  \\
    jackal_hybrid_3   & 2.49                     & 0.9    & \textcolor{green}{2.48 } & 0.89  \\
    jackal_indoor_0   & 4.73                     & 0.16   & \textcolor{green}{4.56 } & 0.19  \\
    jackal_indoor_1   & \textcolor{green}{2.97 } & 0.25   & 3.08                     & 0.26  \\
    jackal_indoor_2   & 4.81                     & 0.39   & \textcolor{green}{4.15 } & 0.65  \\
    jackal_indoor_3   & 3.34                     & 0.15   & \textcolor{green}{3.34 } & 0.13  \\
    jackal_outdoor_0  & 2.51                     & 0.26   & \textcolor{green}{2.49 } & 0.25  \\
    jackal_outdoor_1  & 2.5                      & 0.15   & \textcolor{green}{2.49 } & 0.15  \\
    jackal_outdoor_2  & 2.19                     & 0.05   & \textcolor{green}{2.18 } & 0.06  \\
    \bottomrule
  \end{tabularx}%
  \caption{
      Dense geometric map accuracy (ATE RMSE) with and without PGMO. 
      For more results on PGMO, the interested reader can refer to~\cite{Rosinol21ijrr-Kimera}. 
  }
  \label{tab:pgmo_ablation}
  \vspace{-10mm}
  \end{table}{}


\subsection{Competitor Comparison}
\label{subsec:experiments-competitors}

Since Kimera's original release, other \vislam pipelines have also had updates to improve their performance and capabilities. 
For instance, Vins-Fusion~\cite{Qin19arxiv-VINS-Fusion-odometry} is the successor to the popular Vins-Mono~\cite{Qin18tro-vinsmono}. 
Similarly, ORB-SLAM3~\cite{Campos21-TRO} provides improvements over the successful ORB-SLAM2~\cite{Mur-Artal17tro-ORBSLAM2}.
In this section, we compare Kimera's performance to these pipelines with the latest improvements to Kimera-VIO across several platforms. 
For more comparisons against other open-source pipelines (\eg \cite{Geneva20icra-openVINS,Leutenegger13rss,Blosch15iros}) we refer the reader to~\cite{Rosinol21ijrr-Kimera,Abate23iser-KimeraSelfDriving}. 
As ORB-SLAM3 is a SLAM-only pipeline, we only provide comparisons against ORB-SLAM3 with loop closures enabled in Kimera. 
Because Vins-Fusion cannot do RGB-D VIO, we omit results for the A1 dataset, which uses the D455 camera. 
This is denoted with a dot ($\bullet$) in that region of the table. 
Additionally, as there are no stereo cameras in the CarSim dataset, we show results for Kimera-VIO in monocular mode, and omit results for Vins-Fusion in stereo mode. 

Table~\ref{tab:competitor-vio} compares Kimera-VIO (without loop closures) with Vins-Fusion~\cite{Qin19arxiv-VINS-Fusion-odometry}. 
Kimera is evaluated with external odometry for the Jackal datasets and in monocular mode for the CarSim datasets. 
Kimera is also evaluated using the RGB-D frontend for the A1 dataset. 
Overall, Kimera outperforms Vins-Fusion in the majority of cases, with Vins-Fusion showing failures in several datasets (represented by dashes). 
The exception was in the uHumans2 datasets, where Vins-Fusion in stereo was better by a significant margin. 
The uHumans2 agent does not have any dynamics that would cause disturbances in the IMU data, unlike in the CarSim datasets where car dynamics are simulated and there are frequent accelerations and braking maneuvers. 
CarSim and uHumans2 were both developed in the same simulation environment, so the discrepancy is likely caused by the type of scenes and agent dynamics.


\begin{table}[h!]
  \vspace{-5mm}
  \centering
  \scriptsize
  \begin{tabularx}{\columnwidth}{l *{6}{Y}}
    \toprule
    & \multicolumn{6}{c}{VIO Absolute Translation Error RMSE (No Loop Closures)} \\
    \cmidrule{2-7} & 
      \multicolumn{2}{c}{\textbf{Kimera-VIO}} & 
      \multicolumn{2}{c}{\textbf{Vins-Fusion Mono}} &
      \multicolumn{2}{c}{\textbf{Vins-Fusion Stereo}}
    \\
Dataset & Avg [m] & Std [m] & Avg [m] & Std [m] & Avg [m] & Std [m] \\
  \midrule
  carsim_1  & M \textcolor{green}{0.65}  & 0.03 & -- & -- & $\bullet$ & $\bullet$ \\
  carsim_2  & M \textcolor{green}{1.82}  & 0.09 & -- & -- & $\bullet$ & $\bullet$ \\
  carsim_3  & M \textcolor{green}{2.81}  & 0.32 & -- & -- & $\bullet$ & $\bullet$ \\
  carsim_4  & M \textcolor{green}{3.24}  & 0.22 & -- & -- & $\bullet$ & $\bullet$ \\
  \midrule
  jackal_hybrid_0    & S \textcolor{green}{3.21} & 0.03     & 5.6                      & 0.43     & 3.75                    & 0.04   \\
  jackal_hybrid_1    & S \textcolor{green}{3.73} & 0.48     & 9.71                     & 1.1      & 13.1                    & 15.0   \\
  jackal_hybrid_2    & S \textcolor{green}{8.3 } & 1.09     & 10.57                    & 2.76     & --                      & --     \\
  jackal_hybrid_3    & S \textcolor{green}{11.8} & 7.02     & 41.42                    & 50.36    & 47.8                    & 68.7   \\
  jackal_indoor_0    & S 11.5                    & 1.19     & \textcolor{green}{6.96}  & 1.7      & 7.7                     & 1.67   \\
  jackal_indoor_1    & S 4.48                    & 0.16     & 12.0                     & 7.17     & \textcolor{green}{3.75} & 1.69   \\
  jackal_indoor_2    & S \textcolor{green}{10.7} & 2.45     & --                       & --       & --                      & --     \\
  jackal_indoor_3    & S \textcolor{green}{6.05} & 0.81     & --                       & --       & --                      & --     \\
  jackal_outdoor_0   & S \textcolor{green}{10.6} & 2.0      & --                       & --       & --                      & --     \\
  jackal_outdoor_1   & S \textcolor{green}{12.3} & 1.35     & --                       & --       & 21.6                    & 10.9   \\
  jackal_outdoor_2   & S \textcolor{green}{21.3} & 2.09     & --                       & --       & --                      & --     \\
  \midrule
  a1_simmons_0       & D \textcolor{green}{0.92} & 0.64  & --                       & --       & $\bullet$                 & $\bullet$\\
  \midrule
  uH2_apartment     & S 0.11 & 0.0     & 0.03 & 0.1     & \textcolor{green}{0.01} & 0.01 \\
  uH2_suburb        & S 2.25 & 0.13    & 1.51 & 0.07    & \textcolor{green}{0.23} & 0.03 \\
  uH2_office        & S 0.34 & 0.4     & 0.23 & 0.03    & \textcolor{green}{0.05} & 0.01 \\
  uH2_subway        & S 4.11 & 2.78    & 0.28 & 0.01    & \textcolor{green}{0.16} & 0.01 \\
  \midrule
  MH_01       & S \textcolor{green}{0.10} & 0.03                    & 0.18                     &  0.0                     & 0.26                    & 0.0                     \\
  MH_02       & S 0.10                    & 0.01                    & \textcolor{green}{0.05}  &  0.06                    & 0.2                     & 0.0                     \\
  MH_03       & S \textcolor{green}{0.16} & 0.01                    & --	                     &  --                      & 0.3                     & 0.0                     \\
  MH_04       & S 0.21                    & 3.2                     & \textcolor{green}{0.2 }  &  0.0                     & 0.42                    & 8.41                    \\
  MH_05       & S \textcolor{green}{0.15} & 0.02                    & 0.3                      &  0.01                    & 0.31                    & 0.0                     \\
  V1_01       & S \textcolor{green}{0.05} & 0.02                    & 0.06                     &  0.03                    & 0.11                    & 6.57                    \\
  V1_02       & S \textcolor{green}{0.04} & 0.02                    & 0.27                     &  0.01                    & 0.1                     & 0.00                    \\
  V1_03       & S 0.10                    & 0.02                    & 0.17                     &  0.01                    & \textcolor{green}{0.09} & 0.02                    \\
  V2_01       & S \textcolor{green}{0.06} & 0.09                    & 0.09                     &  0.1                     & 0.14                    & 3.95                    \\
  V2_02       & S \textcolor{green}{0.07} & 0.34                    & --                       &  --                      & 0.12                    & 0.0                     \\
  V2_03       & S 0.19                    & 0.0                     & \textcolor{green}{0.16}  &  0.01                    & 0.33                    & 0.0                     \\
  \bottomrule
\end{tabularx}%
\caption{
    VIO localization accuracy for Kimera compared to Vins-Fusion. 
    Datasets that failed to maintain tracking are noted with dashes. 
    A dot ($\bullet$) denotes that either the pipeline was unable to run on that dataset (\eg no support for RGB-D) or the dataset did not contain relevant sensors (\eg CarSim does not have stereo cameras). 
    The modality used by Kimera is denoted to the left of the first data colum. S = stereo, M = monocular, D = RGB-D. 
}
\label{tab:competitor-vio}
\vspace{-10mm}
\end{table}{}


Table~\ref{tab:competitor-pgo} compares Kimera to Vins-Fusion and ORB-SLAM3, all with loop closures. 
Vins-Fusion and ORB-SLAM3 are evaluated in monocular and stereo/RGB-D mode. 
Note that as ORB-SLAM3 supports RGB-D-Inertial \vislam, we used that configuration for the A1 dataset, however ORB-SLAM3 was unable to maintain consistent tracking in either RGB-D or stereo mode. 
Kimera outperforms its competitors in most cases; Vins-Fusion has the lowest trajectory error in uHumans2  while ORB-SLAM3 mostly exhibits good performance on EuRoC.
In the EuRoC evaluation, ORB-SLAM3 outperformed both Kimera and Vins-Fusion in most cases, with Vins-Fusion outperforming Kimera as well. 
ORB-SLAM3 performs frequent bundle-adjustment optimizations, which undoubtedly lead to better state estimation on smaller datasets like EuRoC. 
However, on longer datasets like in CarSim or the Jackal datasets, the advantage is lessened. 


\begin{table*}[h!]
  \vspace{-4mm}
  \centering
  \scriptsize
  \begin{tabularx}{\textwidth}{l *{10}{Y}}
    \toprule
    & \multicolumn{10}{c}{\vislam Absolute Translation Error RMSE (With Loop Closures)} \\
    \cmidrule{2-11} & 
    \multicolumn{2}{c}{\textbf{Kimera}} & 
    \multicolumn{2}{c}{\textbf{Vins Mono}} & 
    \multicolumn{2}{c}{\textbf{Vins Stereo}} & 
    \multicolumn{2}{c}{\textbf{ORB3 Mono}} & 
    \multicolumn{2}{c}{\textbf{ORB3 S/D}} \\
Dataset & Avg [m] & Std [m] & Avg [m] & Std [m] & Avg [m] & Std [m] & Avg [m] & Std [m] & Avg [m] & Std [m] \\
  \midrule
  carsim_2  & M \textcolor{green}{0.51}  & 0.01     & --   & --                      & $\bullet$                        & $\bullet$          & 22.6 & 4.23   & $\bullet$     & $\bullet$      \\
  carsim_3  & M \textcolor{green}{2.55}  & 0.52     & --   & --                      & $\bullet$                        & $\bullet$          & 61.2 & 20.5   & $\bullet$     & $\bullet$      \\
  carsim_4  & M \textcolor{green}{3.22}  & 0.26     & --   & --                      & $\bullet$                        & $\bullet$          & 15.3 & 6.33   & $\bullet$     & $\bullet$      \\
  \midrule
  jackal_hybrid_0    & S \textcolor{green}{3.21}  & 0.04     & 5.12                           & 1.37                    & 3.85                    & 0.05      & 25.0 & 21.6   & --   & --    \\
  jackal_hybrid_1    & S \textcolor{green}{3.73}  & 0.39     & 6.15                           & 5.39                    & 10.1                    & 14.8      & 44.2 & 9.08   & --   & --    \\
  jackal_hybrid_2    & S \textcolor{green}{8.30}  & 1.09     & 4.47                           & 1.39                    & --                      & --        & --   & --     & --   & --    \\
  jackal_hybrid_3    & S \textcolor{green}{11.8}  & 7.02     & 42.4                           & 52.4                    & 48.1                    & 69.3      & --   & --     & --   & --    \\
  jackal_indoor_0    & S 11.4                     & 1.19     & 6.14                           & 0.91                    & \textcolor{green}{8.01} & 1.89      & --   & --     & 11.9 & 5.03  \\
  jackal_indoor_1    & S 3.98                     & 0.83     & 11.8                           & 8.08                    & \textcolor{green}{3.54} & 1.61      & 16.5 & 6.07   & 5.86 & 5.64  \\
  jackal_indoor_2    & S \textcolor{green}{6.97}  & 2.05     & --                             & --                      & --                      & --        & 72.4 & 0.31   & 15.2 & 5.49  \\
  jackal_indoor_3    & S \textcolor{green}{6.04}  & 0.81     & --                             & --                      & --                      & --        & --   & --     & 23.3 & 14.9  \\
  jackal_outdoor_0   & S \textcolor{green}{10.6}  & 1.99     & --                             & --                      & 20.3                    & 17.3      & --   & --     & 19.8 & 24.8  \\
  jackal_outdoor_1   & S 12.3                     & 1.35     & --                             & --                      & \textcolor{green}{4.53} & 3.46      & 88.9 & 10.8   & 15.7 & 9.38  \\
  jackal_outdoor_2   & S 21.3                     & 2.09     & \textcolor{green}{5.94}        & 2.97                    & 25.3                    & 10.8      & --   & --     & 18.1 & 17.3  \\
  \midrule
  a1_simmons_0       & D \textcolor{green}{0.92}  & 0.64     & --   & --                      &  $\bullet$                       & $\bullet$          & --   & --     & --   & --    \\
  \midrule
  uH2_apartment & S 0.06                     & 0.0      & 0.02 & 0.01                    & \textcolor{green}{0.01} & 0.0       & 0.12 & 0.53   & 0.02 & 0.42  \\
  uH2_suburb    & S 0.69                     & 0.13     & 0.61 & 0.4                     & \textcolor{green}{0.14} & 0.02      & 32.6 & 0.4    & --   & --    \\
  uH2_office    & S 0.11                     & 0.18     & 0.02 & 0.01                    & \textcolor{green}{0.02} & 0.0       & 0.69 & 0.4    & 0.04 & 0.05  \\
  uH2_subway    & S 0.41                     & 2.78     & 0.02 & 0.04                    & \textcolor{green}{0.01} & 0.0       & 0.72 & 0.32   & 0.16 & 0.21  \\
  \midrule
  MH_01         & S 0.09                    & 0.01                    & 0.07                    & 0.01                    & 0.17 & 0.0  & \textcolor{green}{0.05} & 0.0                     & 0.09                    & 0.1                     \\
  MH_02         & S 0.11                    & 0.05                    & \textcolor{green}{0.02} & 0.02                    & 0.15 & 0.01 & 0.06                    & 0.0                     & 0.05                    & 0.0                     \\
  MH_03         & S 0.12                    & 0.01                    & \textcolor{green}{0.04} & 0.05                    & 0.14 & 0.0  & 0.09                    & 0.02                    & 0.06                    & 0.0                     \\
  MH_04         & S 0.15                    & 3.2                     & 0.12                    & 0.01                    & 0.27 & 0.0  & \textcolor{green}{0.08} & 0.02                    & 0.09                    & 0.02                    \\
  MH_05         & S 0.15                    & 0.02                    & 0.13                    & 0.0                     & 0.3  & 0.02 & 0.12                    & 0.01                    & \textcolor{green}{0.08} & 0.02 \\
  V1_01         & S 0.06                    & 0.5                     & \textcolor{green}{0.04} & 0.0                     & 0.08 & 0.0  & 0.06                    & 0.0                     & 0.12                    & 0.01                    \\
  V1_02         & S \textcolor{green}{0.04} & 0.02                    & 0.17                    & 0.0                     & 0.07 & 0.0  & 0.12                    & 0.0                     & 0.06                    & 0.0                     \\
  V1_03         & S 0.10                    & 0.02                    & 0.1                     & 0.01                    & 0.1  & 0.02 & \textcolor{green}{0.08} & 0.0                     & 0.12                    & 0.0                     \\
  V2_01         & S 0.05                    & 0.09                    & 0.08                    & 0.0                     & 0.11 & 0.04 & \textcolor{green}{0.05} & 0.01                    & 0.08                    & 0.0                     \\
  V2_02         & S 0.06                    & 0.34                    & --                      & --                      & 0.09 & 0.0  & 0.09                    & 0.01                    & \textcolor{green}{0.05} & 0.01 \\
  V2_03         & S 0.19                    & 0.0                     & 0.09                    & 0.0                     & 0.1  & 0.02 & 0.14                    & 0.03                    & \textcolor{green}{0.09} & 0.01 \\
  \bottomrule
\end{tabularx}%
\caption{
    \vislam localization accuracy for Kimera compared to Vins-Fusion and ORB-SLAM3. 
    Loop closures are included for all pipelines represented here. 
    Datasets that failed to maintain tracking are noted with dashes. 
    A dot ($\bullet$) denotes that either the pipeline was unable to run on that dataset (\eg no support for RGB-D) or the dataset did not contain relevant sensors (\eg CarSim does not have stereo cameras). 
    The modality used by Kimera is denoted to the left of the first data colum. S = stereo, M = monocular, D = RGB-D. 
}
\label{tab:competitor-pgo}
\vspace{-8mm}
\end{table*}{}





\section{Conclusions}
\label{sec:conclusions}

In this paper, we presented several key improvements to Kimera since its initial release in 2019. 
In particular, we discussed modifications to the Kimera-VIO frontend
    to support additional sensor modalities (\eg monocular, stereo, RGB-D), 
    optional external odometry sources, 
    image feature binning, 
    and an updated keyframe-selection logic. 
We also discussed changes to the backend to include \GNC as an outlier-rejection method for robust pose-graph optimization. 
We provided extensive ablation studies on the impact of these improvements on localization error, across a variety of datasets. 
Additionally, we showcased improvements to the dense 3D mesh reconstruction of Kimera-Semantics with evaluations of Kimera-PGMO on multiple datasets. 
Finally, we demonstrated Kimera's performance as compared to other open-source \vislam pipelines, showing favorable performance, in particular in large-scale datasets. 


%
%
%

\bibliographystyle{IEEEtran}
\bibliography{refs,myRefs}







\end{document}